\documentclass{article} 
\usepackage{iclr2027_conference,times}

\usepackage{amsmath,amsfonts,bm}

\def\eqref#1{equation~\ref{#1}}

\def\1{\bm{1}}

\DeclareMathAlphabet{\mathsfit}{\encodingdefault}{\sfdefault}{m}{sl}
\SetMathAlphabet{\mathsfit}{bold}{\encodingdefault}{\sfdefault}{bx}{n}

\usepackage{hyperref}
\usepackage{url}

\usepackage{dsfont}
\usepackage{bm}
\usepackage{multirow}
\usepackage{adjustbox}
\usepackage{graphicx}
\usepackage{wrapfig}
\usepackage{needspace}
\graphicspath{{figs/}{figures/}} 
\usepackage{amsmath}
\usepackage{float}
\usepackage{caption}
\usepackage{tcolorbox} 
\usepackage{amssymb}
\usepackage{subcaption}
\usepackage{verbatim}
\usepackage{multicol}
\usepackage{booktabs}
\usepackage{makecell} 
\usepackage[normalem]{ulem} 
\usepackage{comment}
\usepackage{soul,color} 
\sethlcolor{yellow}
\newcommand{\highlight}[1]{\hl{#1}}
\usepackage{kotex} 
\usepackage{pifont}
\usepackage{multirow}

\definecolor{markgreen}{RGB}{46, 204, 113}
\definecolor{markred}{RGB}{231, 76, 60}
\definecolor{markorange}{RGB}{243, 156, 18}
\newcommand{\cmark}{{\color{markgreen}\ding{51}}}
\newcommand{\xmark}{{\color{markred}\ding{55}}}
\newcommand{\pmark}{\ensuremath{{\color{markorange}\triangle}}} 
\usepackage[linesnumbered,ruled]{algorithm2e}

\usepackage{colortbl}
\usepackage{xcolor}
\definecolor{citationpurple}{HTML}{B380FF}
\hypersetup{colorlinks=true, citecolor=citationpurple,
            linkcolor=black, urlcolor=black, filecolor=black}
\definecolor{rowgray}{RGB}{240,240,240}
\definecolor{red}{RGB}{255, 102, 102}

\title{Improving Proactive AI Assistance with Hierarchical Procedural Understanding}

\author{
\hspace{-0.06in}
Jin-Seop Lee$^{1}$\quad
TaeYeon Won$^{1}$\quad
SeongJun Jung$^{1}$\quad
\textbf{JungHoon Kim}$^{1}$\quad\\
\textbf{Boyang Albert Li}$^{2}$\quad
\textbf{JinYeong Bak}$^{1}$\quad
\textbf{Jaehong Yoon}$^{2,*}$\quad
\textbf{Jee-Hyong Lee}$^{1,*}$ \\[4pt]
$^{1}$Department of Artificial Intelligence, Sungkyunkwan University, Republic of Korea \\
$^{2}$College of Computing and Data Science, Nanyang Technological University, Singapore \\[3pt]
\textit{$^{*}$Corresponding authors.}
}

\iclrfinalcopy 

\begin{document}

\maketitle
\lhead{Preprint.}

\begin{abstract}
Proactive AI assistants continuously observe a user's activity and decide whether to provide new guidance or remain silent.
To effectively assist users, they should provide appropriate guidance for the task, determine when to provide the next guidance based on the current task progress, and adjust the guidance level to the user's expertise and needs.
Supporting these capabilities requires training and evaluation data that reflect procedural structure and capture how guidance should adapt to task progress and user needs.
However, existing datasets either remain limited to detection-based proactive understanding or provide procedural guidance at a fixed granularity.
Fixed-granularity guidance provides limited information about fine-grained progress and broader procedural context, making it difficult to determine completion in a streaming setting and adapt guidance granularity to the user's needs.
To address these limitations, we introduce the \textit{ProactiveCoach} suite, comprising \textit{ProactiveCoach-Instruct} for training, \textit{ProactiveCoachBench} for evaluation, and fine-tuned vision-language models (VLMs) with an adaptive guidance system.
\textit{ProactiveCoach-Instruct} provides hierarchically structured guidance at the phase, step, and action levels, offering richer supervision for learning task progress and procedural context across different granularities.
\textit{ProactiveCoachBench} evaluates whether models provide appropriate guidance at the right time across different guidance levels and whether they can adapt when the requested guidance level changes.
We fine-tune pretrained VLMs on \textit{ProactiveCoach-Instruct} and demonstrate its effectiveness across different backbones.
Compared with fixed-granularity supervision, hierarchical supervision consistently improves overall performance across various backbones, with gains of up to 9.6\%p.
We further build an adaptive guidance system by combining our fine-tuned model with a lightweight guidance router.
Without additional fine-tuning of the guide model, our system outperforms the in-context adaptation baseline by 57.1\%p in overall performance across four guidance-level transitions.
Our project page is available at \url{https://jinsuby.github.io/ProactiveCoach/}.

\end{abstract}
    
\section{Introduction}
Building AI assistants like JARVIS\footnote{JARVIS stands for ``Just A Rather Very Intelligent System.'' It is a fictional AI assistant used by Tony Stark in the Marvel Cinematic Universe, first appearing in \textit{Iron Man} (2008). It assists Stark by managing information and controlling various systems and devices.} that continuously understand their surroundings and proactively help users when needed has long been a human aspiration.
For example, a future smart-glasses assistant could recognize what a user is currently doing and provide guidance at an appropriate moment, and a future humanoid assistant robot could monitor task progress and proactively provide physical assistance to the user when help is needed~\citep{mmduet2, streamo, em-grade, omnipro, streamready, querystream, proact-vl,response-g1,eyewo}.
A central requirement for realizing such AI assistants is \emph{proactiveness}.

\emph{Proactiveness} refers to continuously observing the user's activity in a streaming setting and deciding whether to provide new guidance or remain silent.
Building on this notion of \emph{proactiveness}, a proactive AI assistant requires three core capabilities.
(1) \emph{It should provide appropriate guidance for the user's task.}
Beyond simply recognizing objects, the assistant should understand the current situation and task progress in activities such as cooking, assembly, and repair, which require a sequence of events to achieve a goal.
(2) \emph{It should provide the next guidance after the currently guided event is completed.}
This requires tracking the progress of the guided event and providing the next guidance when it ends, while remaining silent when no new guidance is needed.
(3) \emph{It should provide guidance at a level appropriate to the user's expertise and needs.}
The assistant should adjust the guidance granularity accordingly and adapt when the requested guidance level changes during an ongoing task.



Although recent studies have introduced training datasets and benchmarks for proactive assistance, existing data do not sufficiently support these core capabilities.
Some datasets remain limited to detection-based proactive understanding, where models detect the occurrence of specific objects or events in a video stream and respond accordingly~\citep{ovobench,streamingbench,proactivevideoqa,omnipro,ipibench}.
Other datasets provide procedural guidance for the user's task, but mainly provide guidance at a fixed granularity~\citep{proassist,streampro,pwr,guideme}.
Since a guided event may consist of multiple finer-grained events, fixed-granularity guidance provides limited information about what has been completed and what remains, making it difficult to track progress and determine completion in a streaming setting.
Since a guided event also belongs to a broader procedure, such guidance provides limited context about its purpose and expected outcome, making it difficult to understand how the current event contributes to the overall procedure.
Fixed-granularity guidance further limits the ability to adjust the guidance level to the user's expertise and needs or adapt when the requested level changes.




\begin{figure}
    \centering
    \includegraphics[width=1\linewidth]{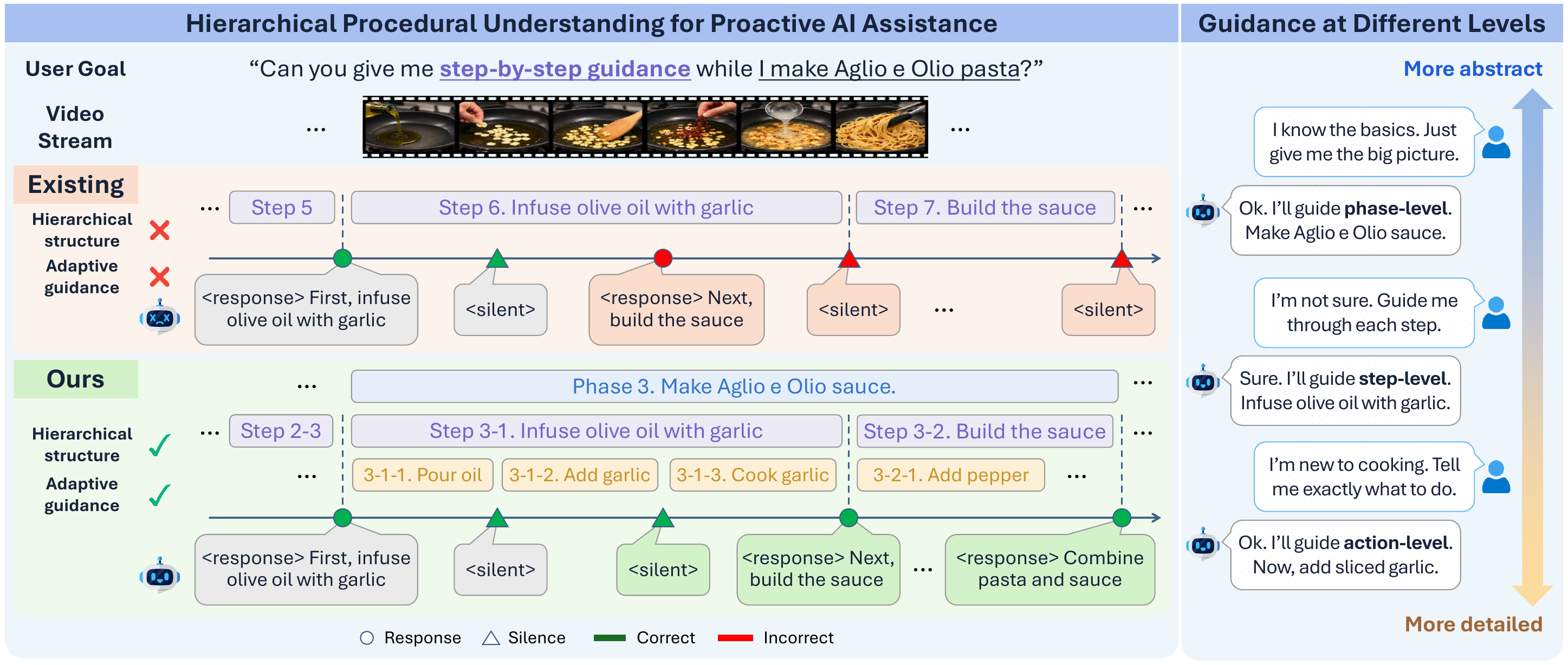}
    \vspace{-0.5cm}
    \caption{Overview of \textsc{ProactiveCoach} suite. It tracks progress and completion across phase, step, and action levels and adapts the guidance level to the user's request.}
    \label{fig:proactive-scenario}
    \vspace{-0.3cm}
\end{figure}

To address these limitations, we introduce the \textit{ProactiveCoach} suite for proactive AI assistance, comprising a training dataset, an evaluation benchmark, and fine-tuned VLMs with an adaptive guidance system.
Our main contributions are as follows.
(1) We introduce \textit{ProactiveCoach-Instruct} for training and \textit{ProactiveCoachBench} for evaluation.
ProactiveCoach-Instruct provides hierarchically structured guidance at the phase, step, and action levels, together with guidance timing, enabling models to learn both fine-grained task progress and higher-level procedural context and to generate guidance at different granularities.
ProactiveCoachBench evaluates whether models provide appropriate guidance at the right time at each guidance level, and further assesses whether they can adapt when the requested guidance level changes during an ongoing task.
(2) We propose \textit{ProactiveCoach}, a hierarchical guidance learning method that fine-tunes pretrained VLMs on ProactiveCoach-Instruct to jointly learn phase-, step-, and action-level proactive guidance in a single model.
This enables different VLM backbones to learn hierarchical procedural understanding and determine what guidance to provide and when to provide it at each level.
(3) We build an adaptive guidance system by combining the fine-tuned VLM with a lightweight guidance router.
The VLM generates multi-level guidance, while the router selects the guidance level requested by the user, allowing the system to dynamically adjust guidance granularity during an ongoing task without additional fine-tuning of the guide model.


Our proposed dataset is constructed using Procedural Data Curation, Procedure Hierarchy Construction, and Proactive Guidance Generation, achieving an overall score of 4.27 in our human quality review.
We also demonstrate that our training data can be effectively applied across different pretrained VLM backbones.
Compared with fixed-granularity supervision, hierarchical supervision consistently improves overall performance across various backbones, with gains of up to 9.6\%p.
Finally, our adaptive guidance system adapts the guidance level when the user's request changes. Without additional fine-tuning of the guide model, it outperforms the in-context adaptation baseline by 57.1\%p in overall performance across four guidance-level transitions.

\section{Related Work}
\begin{table}[t]
\centering
\scriptsize
\setlength{\tabcolsep}{4pt}
\renewcommand{\arraystretch}{1.00}
\begin{adjustbox}{max width=\linewidth}
\begin{tabular}{l%
rrcccc}
\toprule
\multirow{2}{*}{\textbf{Dataset}}
& \multirow{2}{*}{\textbf{\#Videos}}
& \multirow{2}{*}{\textbf{Hours}}
& \multirow{2}{*}{\textbf{Proactive}}
& \textbf{Procedural}
& \textbf{Guidance}
& \textbf{Hierarchical} \\
& & & & \textbf{Tasks} & \textbf{Tasks} & \textbf{Ann.} \\
\midrule

HiREST
& 3.4K & 0.3K
& \xmark & \cmark & \xmark & \cmark \\

GUIDE
& 3.5K & 0.1K
& \xmark & \cmark & \xmark & \cmark \\

Ego4D-HCap
& 4.5K & --
& \xmark & \xmark & \xmark & \cmark \\

Action100M
& 1.2M & 128K
& \xmark & \cmark & \xmark & \cmark \\

\midrule

ProAssist
& 3.4K & 0.5K
& \cmark & \cmark & \cmark & \xmark \\

StreamPro-Bench
& 0.6K & --
& \cmark & \pmark & \pmark & \xmark \\

Pro$^2$Bench
& 29.3K & --
& \cmark & \cmark & \cmark & \xmark \\

GuideMe
& 2.5K & 0.2K
& \cmark & \cmark & \cmark & \xmark \\

\midrule

\textbf{Ours}
& \textbf{7.2K} & \textbf{1.5K}
& \cmark & \cmark & \cmark & \cmark \\

\bottomrule
\end{tabular}
\end{adjustbox}
\vspace{-0.1cm}
\caption{Comparison of existing datasets for hierarchical video understanding and proactive procedural assistance. \pmark{} denotes partial task coverage.}
\vspace{-0.4cm}
\label{tab:benchmark-comparison}
\end{table}

\paragraph{Hierarchical Video Understanding.}
Hierarchical video understanding aims to understand video content at multiple levels, from fine-grained actions to higher-level activities~\citep{goalstep,videorecap,openhsvu}.
HiREST~\citep{hirest} connects video and moment retrieval with step-by-step descriptions, while GUIDE~\citep{guide} identifies common procedures across multiple videos and links them to the detailed steps in each video.
Ego4D-HCap~\citep{videorecap} recursively generates captions at multiple levels, from short action descriptions to long-form video summaries, while Action100M~\citep{action100m} hierarchically segments videos and provides action and context annotations for segments of different durations.
However, hierarchical descriptions cannot be directly converted into guidance.
In proactive procedural assistance, the assistant must identify goal-relevant activities and determine what guidance to provide, when to provide it, and at what level of detail.


\paragraph{Proactive Procedural Assistance.}
Unlike reactive AI, which typically responds to explicit user queries, proactive AI determines not only \textit{what to say} but also whether and \textit{when to respond} based on a continuous stream of observations. Some studies remain limited to detection-based proactive understanding, where models detect specific objects or events in a video stream and respond accordingly~\citep{ovobench,streamingbench,proactivevideoqa,omnipro,ipibench}.
More recently, proactive AI assistance has moved beyond such detection-based settings, requiring models to determine what guidance is needed and when to provide it based on the user's task goal and ongoing task context~\citep{proassist,streampro,pwr,guideme}.

ProAssist~\citep{proassist} and StreamPro-Bench~\citep{streampro} focus on providing proactive step-level guidance according to task progress.
Pro$^{2}$Bench~\citep{pwr} further considers recognizing when the user's behavior does not follow the intended procedure and providing recovery guidance, while GuideMe~\citep{guideme} covers task completion verification, error detection, and corrective guidance.
However, existing datasets and benchmarks primarily focus on step-level guidance, providing limited supervision and evaluation for understanding progress and completion across procedural levels.
They also provide limited evaluation of whether models can provide guidance across different procedural levels and adapt when the requested guidance level changes.

\section{ProactiveCoach-Instruct \& ProactiveCoachBench}
\label{hierarchical understanding for proactive procedural assistance}
In this section, we introduce ProactiveCoach-Instruct, our training dataset, and ProactiveCoachBench, our evaluation benchmark.
We describe their design and statistics in Sec.~\ref{dataset overview}, 
present the construction pipeline in Sec.~\ref{construction pipeline}, and present quality review in Sec.~\ref{quality review}.

\begin{figure}[t]
    \centering
    \includegraphics[width=\linewidth]{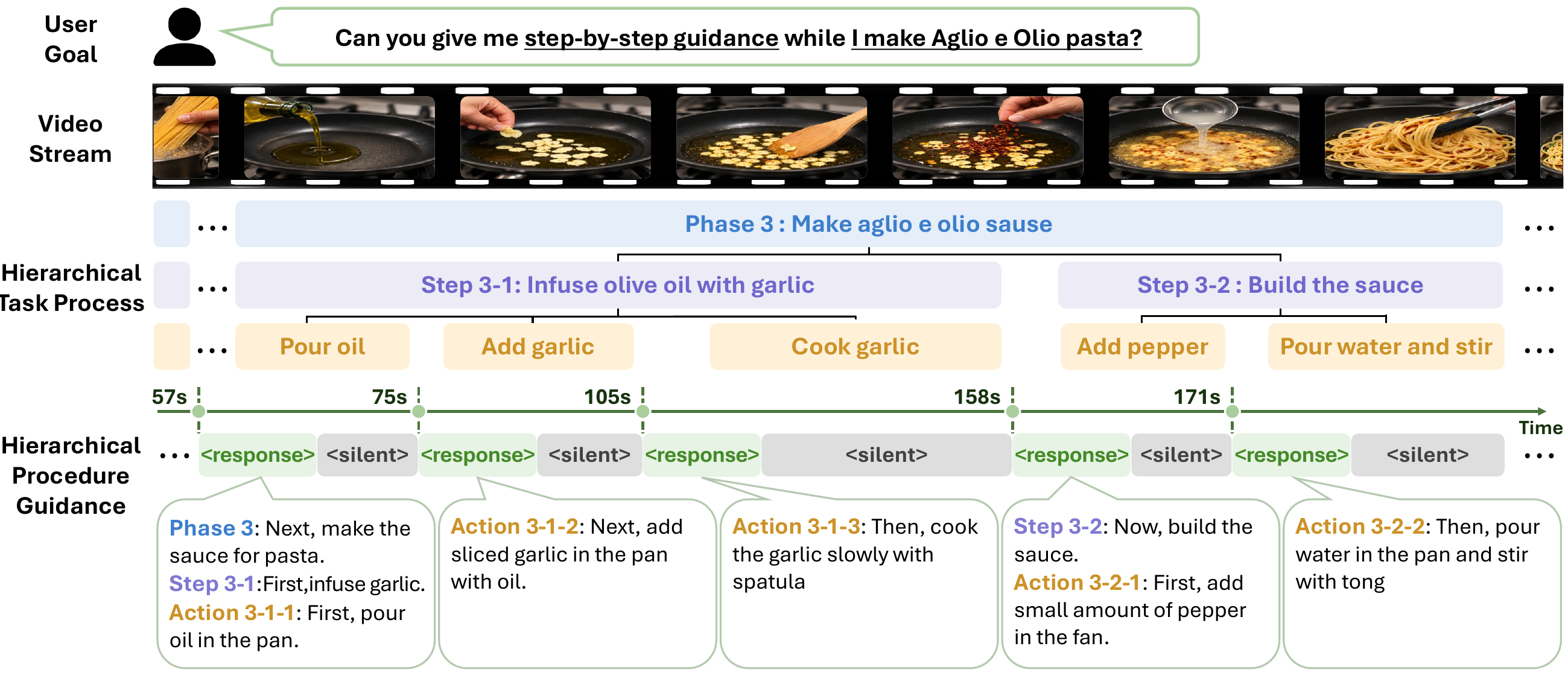}
    \vspace{-0.4cm}
    \caption{Overview of hierarchically structured guidance in our proposed dataset.}
    \label{fig:hipropa-example}
    \vspace{-0.2cm}
\end{figure}

\begin{figure}[t]
    \centering
    \captionsetup{font=small,skip=3pt}
    \begin{minipage}[t]{0.48\linewidth}
        \centering
        \vspace{0pt}
        \includegraphics[width=0.74\linewidth]{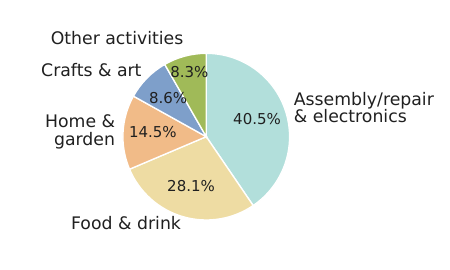}
        \caption{Domain distribution of our dataset.}
        \vspace{-0.2cm}
        \label{fig:hipropa-statistics}
        \label{fig:hipropa-domain-panel}
    \end{minipage}\hfill
    \begin{minipage}[t]{0.48\linewidth}
        \centering
        \vspace{0pt}
        \includegraphics[width=0.74\linewidth]{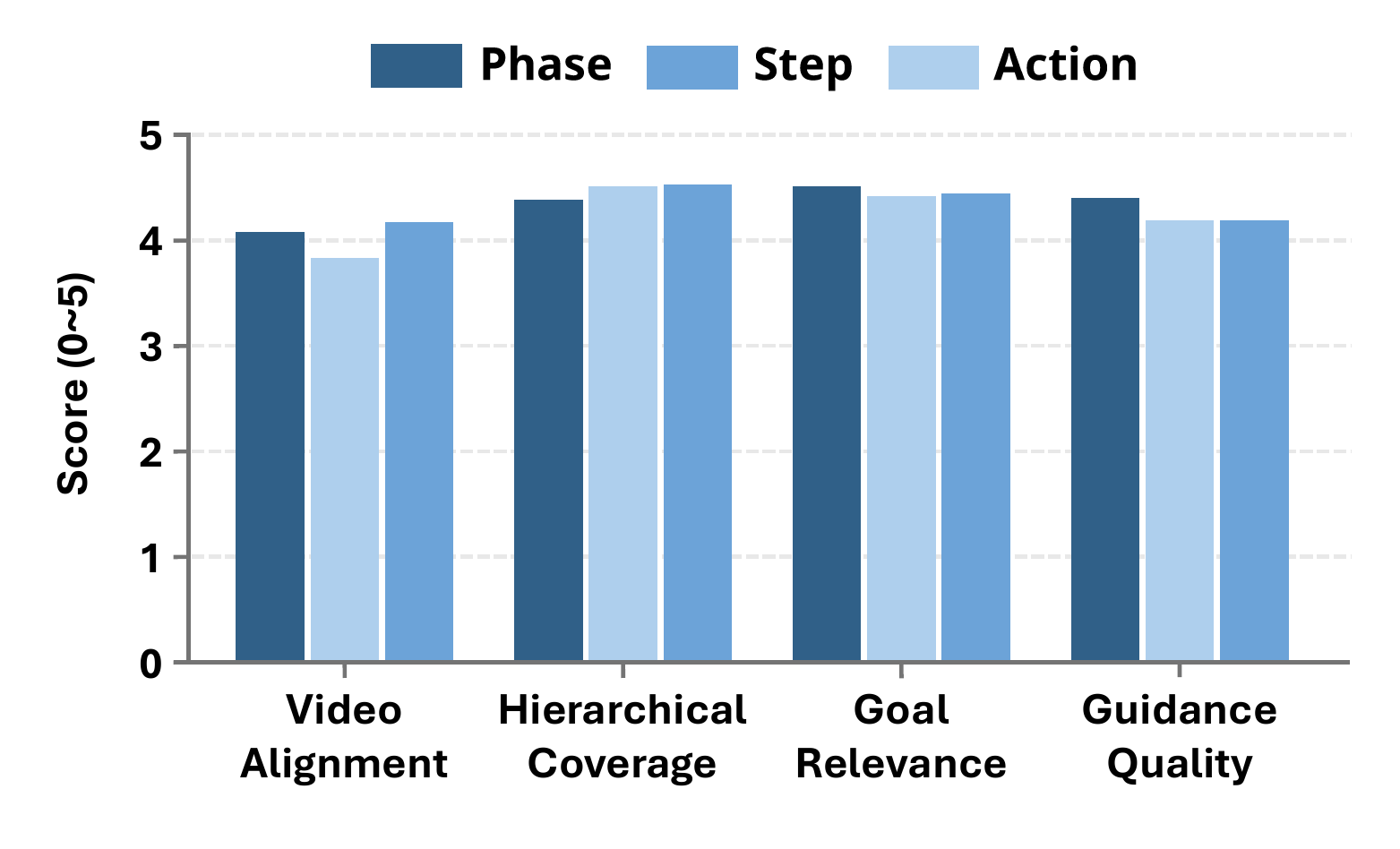}
        \caption{Human quality review of our dataset.}
        \vspace{-0.2cm}
        \label{fig:hipropa-quality}
    \end{minipage}
\end{figure}

\subsection{Dataset Overview}
\vspace{-0.2cm}
\label{dataset overview}
To support hierarchical procedural understanding for proactive procedural assistance, we structure goal-relevant guidance into three levels: phase, step, and action.
This hierarchy allows lower-level progress to provide evidence for the completion of guided events, while higher-level context helps clarify their purpose and expected outcome.
Based on this design, we construct \textsc{ProactiveCoach-Instruct} and \textsc{ProactiveCoachBench} with hierarchically structured guidance and guidance timing at each level, supporting learning and evaluation across procedural levels. This hierarchical structure also supports guidance at different granularities according to the user's needs.

Figure~\ref{fig:hipropa-example} illustrates the hierarchically structured guidance in ProactiveCoach-Instruct and ProactiveCoachBench.
A phase is a high-level unit that achieves an intermediate goal through multiple steps, while a step is a procedural unit that achieves a specific outcome through multiple actions.
An action is an observable fine-grained event that can be guided independently and serves as the finest guidance level.
Each sample consists of a task goal, a video, and hierarchical guidance annotations, with each guidance sentence linked to its event span and guidance time.
Guidance timing is defined independently at each level. The next guidance is provided immediately after the previously guided event at that level ends.
For each video chunk, we combine multi-level guidance in phase, step, and action order into a single \texttt{<response>}.
If no new guidance is required at any level, the chunk is labeled \texttt{<silent>}.

The data contain 11,119 samples constructed from 7,232 first-person videos across eight source datasets: Ego4D Goal-Step~\citep{goalstep}, Ego-Exo4D~\citep{ego-exo4d}, EPIC-KITCHENS~\citep{epic-kitchens}, CaptainCook4D~\citep{captaincook4d}, EgoExoLearn~\citep{egoexolearn}, HD-EPIC~\citep{hd-epic}, WTaG~\citep{wtag}, and HoloAssist~\citep{holoassist}.
As shown in Figure~\ref{fig:hipropa-statistics}, the data cover diverse domains, including assembly, repair, and electronics; food and beverage preparation; and household and gardening activities.
We use 10,007 samples as ProactiveCoach-Instruct for training and 1,112 samples as ProactiveCoachBench for evaluation.
On average, each sequence contains 4.14 phases, with 3.35 steps per phase and 2.47 actions per step.
Further dataset statistics are provided in Appendix~\ref{more details of huppa}.

\begin{figure}[t]
    \centering
    \includegraphics[width=\linewidth]{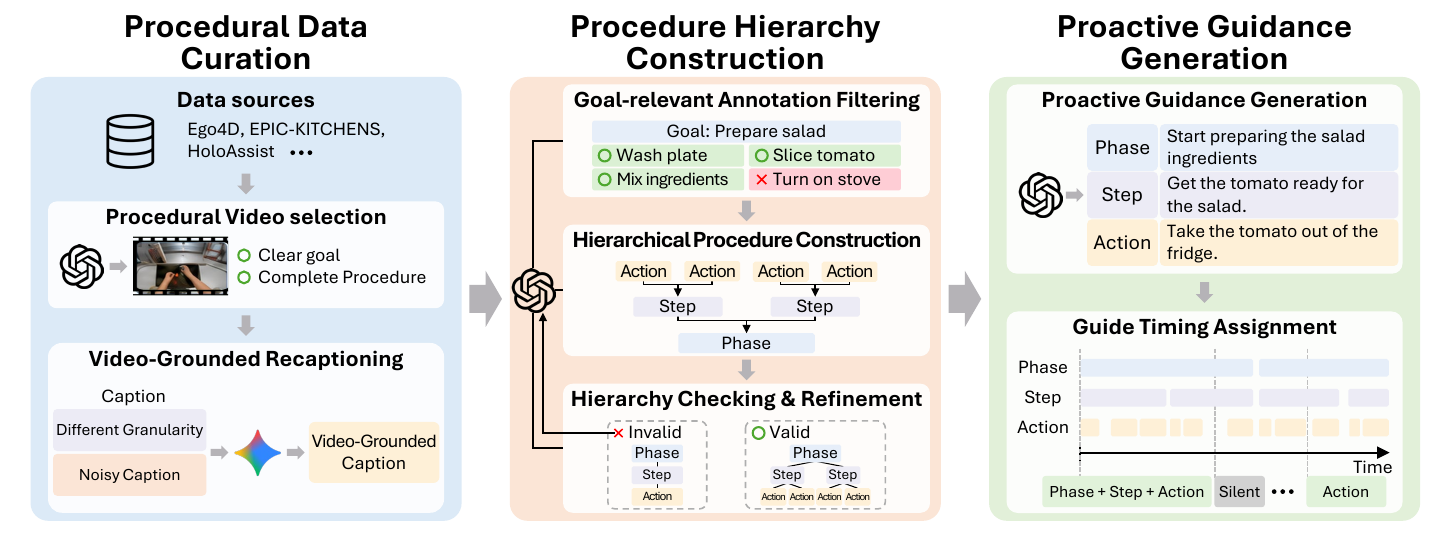}
    \vspace{-0.4cm}
    \caption{Construction pipeline for our proposed dataset.}
    \vspace{-0.3cm}
    \label{fig:dataset-construction}
\end{figure}

\subsection{Construction Pipeline}
\vspace{-0.2cm}
\label{construction pipeline}
The proposed dataset is constructed through three stages: procedural data curation, procedure hierarchy construction, and proactive guidance generation, as shown in Figure~\ref{fig:dataset-construction}.


\vspace{-0.2cm}
\paragraph{Procedural Data Curation.}
We filter goal-oriented procedural videos from eight egocentric video datasets.
The source datasets may contain videos that are unsuitable for procedural assistance, such as those with unclear task goals or insufficient coverage of key events.
Therefore, in procedural video selection, we use GPT-4.1-mini to assess the clarity of the task goal and the coverage of key events based on the existing annotations, and retain videos suitable for procedural assistance.
In addition, the source datasets contain annotations with different levels of granularity, inconsistencies with the video, or missing relevant events.
To address these issues, we perform video-grounded recaptioning using Gemini 2.5 Flash with the selected videos and their existing annotations.
The model revises captions that do not match the video, adds missing captions, and aligns each caption with its corresponding temporal segment in the video.


\vspace{-0.2cm}
\paragraph{Procedure Hierarchy Construction.}
The recaptioned annotations may still include content that is not necessary for accomplishing the task goal.
Therefore, in goal-relevant annotation filtering, we use GPT-4.1-mini to retain only the annotations necessary for achieving the task goal.
Then, we perform hierarchical procedure construction to organize the selected annotations into a procedural hierarchy.
First, we consolidate the selected annotations into actions, each representing an independent unit of execution.
We then group multiple actions into a step representing a procedural unit, and multiple steps into a phase representing an intermediate goal, forming a phase–step–action hierarchy.
The temporal span of each action is defined by its observed start and end times in the video.
The temporal spans of each step and phase are then defined to cover their child actions and steps, respectively.
This produces a hierarchical procedure structure that serves as the basis for hierarchically structured guidance.
However, this process may produce redundant structures where different hierarchical levels describe the same execution.
We therefore perform hierarchy checking and refinement to eliminate these redundancies and ensure hierarchical consistency.
Using GPT-4.1-mini, we check whether each step is properly composed of its actions, each phase is properly composed of its steps, and their temporal spans are consistent.
Based on this check, we reconstruct the hierarchy and repeat the refinement process until the structure is consistent.


\vspace{-0.2cm}
\paragraph{Proactive Guidance Generation.}
To construct hierarchically structured guidance, we perform level-specific guidance generation.
Given the constructed phase–step–action hierarchy and the user's goal information at each level, we  generate proactive guidance for the phase, step, and action levels by using GPT-4.1-mini.
Then, we perform guide timing assignment to align the generated guidance with task progress based on the temporal spans at each level.
At each level, the next guidance is assigned to the frame immediately after the current guided event ends.
If the current guided event ends at only one level, we provide the next guidance only for that level.
If guided events end at multiple levels at the same time, we provide the next guidance for all corresponding levels together.
If no guided event ends at a given frame, we label that frame as silent.
Through this process, we construct \textit{ProactiveCoach-Instruct} and \textit{ProactiveCoachBench} with hierarchically structured guidance for proactive AI assistance.



\vspace{-0.1cm}
\subsection{Quality Review}
\label{quality review}
\vspace{-0.1cm}
We conduct a human evaluation to assess whether the annotations accurately reflect the procedural structure of the video and provide appropriate guidance.
We sample 50 examples from ProactiveCoachBench and evaluate them at the phase, step, and action levels using four criteria: \textit{video alignment}, \textit{hierarchical coverage}, \textit{goal relevance}, and \textit{guidance quality}.
\textit{Video alignment} measures whether the annotated content, temporal spans, and guidance timings are consistent with the observed video.
\textit{Hierarchical coverage} evaluates whether the lower-level executions sufficiently cover the content required by the higher-level procedure, while \textit{goal relevance} measures whether the generated guidance contributes to achieving the task goal.
\textit{Guidance quality} evaluates whether each guidance sentence clearly and accurately describes what the user should do at the corresponding level.
As shown in Figure~\ref{fig:hipropa-quality}, the annotations achieve scores above 4.0 on most evaluation criteria, with an overall score of 4.27.
These results indicate that the annotations reliably capture video-grounded procedural hierarchies and provide goal-relevant guidance at each level.
Details of quality review are provided in Appendix~\ref{details of quality review}.



\section{ProactiveCoach}
\vspace{-0.1cm}
\label{learning from huppa}
We propose ProactiveCoach, a hierarchical guidance learning method that jointly learns phase-, step-, and action-level proactive guidance in a single vision-language model (VLM).
Using ProactiveCoach-Instruct, we train the model to learn hierarchical procedural understanding across the three levels.
For supervised fine-tuning (SFT), we organize guidance across the three levels into a single output sequence that preserves their hierarchical relationships and guidance timing.
Each guidance sentence is paired with its level and a hierarchical ID that encodes the parent--child structure.
This makes the relationship between lower-level events and higher-level procedural units explicit in the training targets.

Let $I$ denote the instruction specifying the task goal, $V_t^{w}$ the video frames from up to $w$ consecutive chunks ending at the current chunk $t$, and $y_t$ the ground-truth output for that chunk.
We denote the ground-truth output history by $H_t^{\mathrm{GT}}=(y_1,\ldots,y_{t-1})$.
Given $I$, $V_t^{w}$, and $H_t^{\mathrm{GT}}$, the model is trained to output \texttt{<silent>} when no new guidance is needed and \texttt{<response>} followed by the corresponding guidance otherwise.
If new guidance is needed at one level, the response contains guidance for that level.
If new guidance is needed at multiple levels, the corresponding guidance is combined into a single response.
During training, we use teacher forcing and provide previous ground-truth outputs as the output history.

In procedural videos, \texttt{<silent>} labels are more frequent than \texttt{<response>} labels.
To prevent repeated silence from dominating the loss, we focus supervision on transitions between response and silence, following a prior work~\citep{videochat3}.
Specifically, we select all \texttt{<response>} chunks and each \texttt{<silent>} chunk immediately following a \texttt{<response>} chunk for supervision.
We then randomly sample additional \texttt{<silent>} chunks from the remaining chunks, matching the total number of chunks selected above.
We compute the loss only on output tokens selected for supervision.
The objective is as follows:
\begin{equation}
\mathcal{L}_{\mathrm{SFT}}(\theta)
=
-\frac{1}{M}
\sum_{t=1}^{T}
m_t
\log p_\theta
\left(
y_t
\mid
I,V_t^{w},H_t^{\mathrm{GT}}
\right).
\label{eq:hipropa-sft}
\end{equation}
Here, $T$ is the total number of chunks, and $m_t\in{0,1}$ indicates whether chunk $t$ is selected for supervision.
We normalize the loss by $M=\sum_{t=1}^{T}m_t$, the total number of selected chunks.
Training on hierarchically structured guidance encourages the model to understand the relationships between finer-grained events and broader procedural context. Lower-level events provide evidence about what has been completed and what remains, while higher-level context helps clarify the purpose and expected outcome of the current guided event. Together with the observed video, this hierarchical understanding provides a stronger basis for deciding what guidance to provide and when to provide it at each level.

\section{Experiments}
\label{sec:experiments}
\vspace{-0.2cm}
\subsection{Experimental Setup}
\label{sec:experimental-setup}
\vspace{-0.1cm}
\paragraph{Datasets and Baselines.}
We evaluate on ProactiveCoachBench and the publicly available validation split of EgoProactive, an existing benchmark for proactive procedural assistance.
We evaluate guidance performance at the phase, step, and action levels on ProactiveCoachBench, and at the step level on EgoProactive~\citep{pwr}.
We compare against four groups of models: Proprietary Frontier Models~\citep{gemini2.5,claudehaiku,gpt5}, Multimodal Models~\citep{qwen3-vl,qwen3.5,videochat-r1.5,internvl3,videollama3,llava-video}, Guidance Assistant Models~\citep{proassist}, and Streaming Models~\citep{videollm-online,mmduet2,videochat3}.
For guidance assistant models, we use ProAssist, whose model weights are publicly available, and for streaming models, we use VideoLLM-Online, MMDuet2, and VideoChat3-4B as baselines.
The details of baselines are provided in Appendix~\ref{evaluation inputs and baseline prompts}.


\paragraph{Evaluation Protocol.}
Following prior evaluation protocols~\citep{pwr,guideme}, we evaluate models at predefined anchor points.
These anchor points are categorized into \textit{Response Anchors} and \textit{Silent Anchors}.
At \textit{Response Anchors}, the model is expected to generate appropriate guidance, whereas at \textit{Silent Anchors}, it is expected to remain silent and avoid providing unnecessary guidance.
For ProactiveCoachBench, we use all ground-truth guidance timestamps as \textit{Response Anchors}.
For \textit{Silent Anchors}, we randomly sample a timestamp with no ground-truth guidance from the central 30--70\% of each phase, step, or action segment.
For EgoProactive, we use the original evaluation anchor points.
At each anchor, all models receive the most recent 10 seconds of video preceding the anchor, together with the complete ground-truth dialogue history prior to that point.
We use ground-truth dialogue history rather than model-generated responses to prevent previous incorrect or missed guidance from affecting subsequent decisions, enabling a fair comparison across models under the same dialogue history.
This protocol therefore evaluates both whether the model correctly decides when to provide guidance and whether the provided guidance is appropriate.


\paragraph{Metrics.}
We use sF1, gF1, and PQS to evaluate proactive procedural assistance, following prior work~\citep{pwr,guideme}.

sF1 jointly evaluates how well the generated guidance matches the ground-truth guidance and whether the model provides all necessary guidance.
We first match ground-truth and generated guidance one-to-one based on both timing and content, and sum the semantic similarities of the matched pairs.
Then, we compute sPrecision and sRecall by dividing this sum by the numbers of generated and ground-truth guidance instances, respectively.
sF1 is calculated as the harmonic mean of sPrecision and sRecall~\citep{guideme}.

gF1 evaluates whether the model correctly decides when to provide guidance and when to remain silent.
It considers only whether the model responds or remains silent at each anchor point and does not consider the guidance content.
We compute precision and recall separately for response and silence, obtain the corresponding F1 scores, and calculate gF1 as their geometric mean~\citep{pwr}.

PQS jointly evaluates response--silence decisions and the quality of the provided guidance.
At a silent anchor, the model receives a score if it remains silent and receives zero otherwise.
At a response anchor, the model receives a score based on the relevance, specificity, actionability, and conciseness of the generated guidance, and receives zero if it remains silent.
PQS is computed by averaging these scores across all anchor points~\citep{pwr}.
We use GPT-5.2 to evaluate guidance content quality.
Details of the metrics are provided in Appendix~\ref{details of metrics}.


\paragraph{Implementation Details.}
We use Qwen3-VL-4B/8B and Qwen3.5-4B/9B as backbone models and train for three epochs using eight B200 GPUs.
Videos are sampled at 1 FPS and grouped into 2-second chunks, with the model deciding whether to respond at each chunk.
We apply a causal block-sparse attention mask that restricts each token to frames from the current and preceding 10 seconds, while preserving access to the ground-truth instruction and all previous response and silence tokens.
Detailed training configurations are provided in Appendix~\ref{training configurations}.


\providecommand{\TSnoft}{}
\providecommand{\TSyesft}{\ensuremath{\checkmark}}
\begin{table*}[t]
  \definecolor{TSoursblue}{RGB}{225,244,248}
  \renewcommand{\cellcolor}[1]{} 
  \centering
  \scriptsize
  \setlength{\tabcolsep}{2.0pt}
  \renewcommand{\arraystretch}{1.30}
  \begin{adjustbox}{max width=\textwidth}
  \begin{tabular}{l|c|ccc|c|ccc|c|ccc|c|ccc|c}
    \noalign{\hrule height 1pt}
    \multirow{3}{*}{\textbf{Model}} & \multirow{3}{*}{\textbf{Ours}} & \multicolumn{12}{c|}{\textbf{ProactiveCoachBench}} & \multicolumn{4}{c}{\multirow{2}{*}{\textbf{EgoProactive}}} \\
    \cline{3-14}
    & & \multicolumn{4}{c|}{Phase} & \multicolumn{4}{c|}{Step} & \multicolumn{4}{c|}{Action} & \multicolumn{4}{c}{} \\
    \cline{3-6}\cline{7-10}\cline{11-14}\cline{15-18}
    & & sF1 & gF1 & PQS & Avg.$\uparrow$ & sF1 & gF1 & PQS & Avg.$\uparrow$ & sF1 & gF1 & PQS & Avg.$\uparrow$ & sF1 & gF1 & PQS & Avg.$\uparrow$ \\
    \noalign{\hrule height 0.6pt}
    \multicolumn{18}{c}{Proprietary Frontier Models}  \\ \hline
    Gemini 3 Flash Preview & \TSnoft & \cellcolor{gray!15}43.01 & \cellcolor{gray!15}70.36 & \cellcolor{gray!15}52.87 & \cellcolor{gray!15}55.41 & \cellcolor{gray!15}41.15 & \cellcolor{gray!15}59.00 & \cellcolor{gray!15}41.88 & \cellcolor{gray!15}47.34 & \cellcolor{gray!15}45.08 & \cellcolor{gray!15}57.45 & \cellcolor{gray!15}36.88 & \cellcolor{gray!15}46.47 & \cellcolor{gray!15}38.62 & \cellcolor{gray!15}64.69 & \cellcolor{gray!15}50.73 & \cellcolor{gray!15}51.35 \\ 
    Claude Haiku 4.5 & \TSnoft & \cellcolor{gray!15}43.25 & \cellcolor{gray!15}27.70 & \cellcolor{gray!15}18.18 & \cellcolor{gray!15}29.71 & \cellcolor{gray!15}41.51 & \cellcolor{gray!15}22.54 & \cellcolor{gray!15}22.89 & \cellcolor{gray!15}28.98 & \cellcolor{gray!15}41.60 & \cellcolor{gray!15}23.90 & \cellcolor{gray!15}19.80 & \cellcolor{gray!15}28.43 & \cellcolor{gray!15}38.81 & \cellcolor{gray!15}52.63 & \cellcolor{gray!15}34.20 & \cellcolor{gray!15}41.88 \\ 
    GPT‑5.4 mini & \TSnoft & \cellcolor{gray!15}33.66 & \cellcolor{gray!15}50.03 & \cellcolor{gray!15}31.23 & \cellcolor{gray!15}38.31 & \cellcolor{gray!15}32.47 & \cellcolor{gray!15}51.21 & \cellcolor{gray!15}34.47 & \cellcolor{gray!15}39.38 & \cellcolor{gray!15}34.47 & \cellcolor{gray!15}53.87 & \cellcolor{gray!15}35.82 & \cellcolor{gray!15}41.39 & \cellcolor{gray!15}39.71 & \cellcolor{gray!15}53.60 & \cellcolor{gray!15}36.18 & \cellcolor{gray!15}43.17 \\ 
    \noalign{\hrule height 0.6pt}
    \multicolumn{18}{c}{Guidance Assist Models} \\ \hline
    ProAssist-8B & \TSnoft & \cellcolor{gray!15}11.06 & \cellcolor{gray!15}38.86 & \cellcolor{gray!15}49.59 & \cellcolor{gray!15}33.17 & \cellcolor{gray!15}9.45 & \cellcolor{gray!15}33.72 & \cellcolor{gray!15}49.03 & \cellcolor{gray!15}30.73 & \cellcolor{gray!15}9.83 & \cellcolor{gray!15}30.96 & \cellcolor{gray!15}48.45 & \cellcolor{gray!15}29.75 & \cellcolor{gray!15}6.07 & \cellcolor{gray!15}24.64 & \cellcolor{gray!15}45.74 & \cellcolor{gray!15}25.49 \\ 
    \noalign{\hrule height 0.6pt}
    \multicolumn{18}{c}{Streaming Models} \\ \hline
    VideoLLM-Online & \TSnoft & \cellcolor{gray!15}0.45 & \cellcolor{gray!15}8.97 & \cellcolor{gray!15}49.97 & \cellcolor{gray!15}19.80 & \cellcolor{gray!15}0.32 & \cellcolor{gray!15}6.48 & \cellcolor{gray!15}49.95 & \cellcolor{gray!15}18.92 & \cellcolor{gray!15}0.34 & \cellcolor{gray!15}5.99 & \cellcolor{gray!15}49.98 & \cellcolor{gray!15}18.77 & \cellcolor{gray!15}0.52 & \cellcolor{gray!15}9.44 & \cellcolor{gray!15}46.17 & \cellcolor{gray!15}18.71 \\ 
    MMDuet2 & \TSnoft & \cellcolor{gray!15}9.97 & \cellcolor{gray!15}40.41 & \cellcolor{gray!15}41.98 & \cellcolor{gray!15}30.79 & \cellcolor{gray!15}19.96 & \cellcolor{gray!15}48.25 & \cellcolor{gray!15}37.93 & \cellcolor{gray!15}35.38 & \cellcolor{gray!15}19.80 & \cellcolor{gray!15}49.25 & \cellcolor{gray!15}36.38 & \cellcolor{gray!15}35.14 & \cellcolor{gray!15}30.63 & \cellcolor{gray!15}44.71 & \cellcolor{gray!15}27.53 & \cellcolor{gray!15}34.29 \\ 
    VideoChat3-4B & \TSnoft & \cellcolor{gray!15}35.91 & \cellcolor{gray!15}19.12 & \cellcolor{gray!15}14.38 & \cellcolor{gray!15}23.14 & \cellcolor{gray!15}35.12 & \cellcolor{gray!15}25.63 & \cellcolor{gray!15}19.85 & \cellcolor{gray!15}26.87 & \cellcolor{gray!15}37.83 & \cellcolor{gray!15}18.00 & \cellcolor{gray!15}16.64 & \cellcolor{gray!15}24.16 & \cellcolor{gray!15}39.46 & \cellcolor{gray!15}36.32 & \cellcolor{gray!15}24.21 & \cellcolor{gray!15}33.33 \\ 
    \noalign{\hrule height 0.6pt}
    \multicolumn{18}{c}{Multimodal Models}
    \\ \hline
    VideoChat-R1.5-7B & \TSnoft & \cellcolor{gray!15}0.05 & \cellcolor{gray!15}2.41 & \cellcolor{gray!15}49.81 & \cellcolor{gray!15}17.42 & \cellcolor{gray!15}0.07 & \cellcolor{gray!15}3.93 & \cellcolor{gray!15}49.72 & \cellcolor{gray!15}17.91 & \cellcolor{gray!15}0.11 & \cellcolor{gray!15}3.33 & \cellcolor{gray!15}49.43 & \cellcolor{gray!15}17.62 & \cellcolor{gray!15}0.20 & \cellcolor{gray!15}4.67 & \cellcolor{gray!15}45.69 & \cellcolor{gray!15}16.85 \\ 
    InternVL3-8B & \TSnoft & \cellcolor{gray!15}13.86 & \cellcolor{gray!15}45.11 & \cellcolor{gray!15}41.13 & \cellcolor{gray!15}33.37 & \cellcolor{gray!15}25.12 & \cellcolor{gray!15}52.14 & \cellcolor{gray!15}38.96 & \cellcolor{gray!15}38.74 & \cellcolor{gray!15}34.71 & \cellcolor{gray!15}54.90 & \cellcolor{gray!15}34.36 & \cellcolor{gray!15}41.32 & \cellcolor{gray!15}18.04 & \cellcolor{gray!15}38.08 & \cellcolor{gray!15}43.38 & \cellcolor{gray!15}33.17 \\ 
    VideoLLaMA3-7B & \TSnoft & \cellcolor{gray!15}15.90 & \cellcolor{gray!15}44.46 & \cellcolor{gray!15}35.63 & \cellcolor{gray!15}31.99 & \cellcolor{gray!15}21.66 & \cellcolor{gray!15}48.82 & \cellcolor{gray!15}38.15 & \cellcolor{gray!15}36.21 & \cellcolor{gray!15}27.74 & \cellcolor{gray!15}50.08 & \cellcolor{gray!15}34.16 & \cellcolor{gray!15}37.33 & \cellcolor{gray!15}1.07 & \cellcolor{gray!15}9.14 & \cellcolor{gray!15}41.80 & \cellcolor{gray!15}17.34 \\ 
    LLaVA-Video-7B & \TSnoft & \cellcolor{gray!15}20.34 & \cellcolor{gray!15}47.74 & \cellcolor{gray!15}35.26 & \cellcolor{gray!15}34.45 & \cellcolor{gray!15}30.19 & \cellcolor{gray!15}51.65 & \cellcolor{gray!15}34.90 & \cellcolor{gray!15}38.92 & \cellcolor{gray!15}25.34 & \cellcolor{gray!15}51.55 & \cellcolor{gray!15}40.10 & \cellcolor{gray!15}39.00 & \cellcolor{gray!15}35.73 & \cellcolor{gray!15}47.65 & \cellcolor{gray!15}33.05 & \cellcolor{gray!15}38.81 \\ 
    Qwen3-VL-32B & \TSnoft & \cellcolor{gray!15}41.56 & \cellcolor{gray!15}38.50 & \cellcolor{gray!15}22.11 & \cellcolor{gray!15}34.06 & \cellcolor{gray!15}38.46 & \cellcolor{gray!15}23.89 & \cellcolor{gray!15}21.71 & \cellcolor{gray!15}28.02 & \cellcolor{gray!15}38.85 & \cellcolor{gray!15}20.84 & \cellcolor{gray!15}17.25 & \cellcolor{gray!15}25.65 & \cellcolor{gray!15}32.06 & \cellcolor{gray!15}49.81 & \cellcolor{gray!15}40.14 & \cellcolor{gray!15}40.67 \\ 
    \noalign{\hrule height 0.6pt}
    \rowcolor{TSoursblue}
    {} & \TSnoft & \cellcolor{gray!15}35.93 & \cellcolor{gray!15}25.82 & \cellcolor{gray!15}18.19 & \cellcolor{gray!15}26.65 & \cellcolor{gray!15}36.50 & \cellcolor{gray!15}26.01 & \cellcolor{gray!15}20.99 & \cellcolor{gray!15}27.83 & \cellcolor{gray!15}37.46 & \cellcolor{gray!15}27.78 & \cellcolor{gray!15}18.65 & \cellcolor{gray!15}27.96 & \cellcolor{gray!15}\textbf{39.93} & \cellcolor{gray!15}35.90 & \cellcolor{gray!15}24.33 & \cellcolor{gray!15}33.39 \\ 
    \rowcolor{TSoursblue}
    \multirow{-2}{*}{Qwen3-VL-4B} & \TSyesft & \cellcolor{gray!15}\textbf{53.15} & \cellcolor{gray!15}\textbf{74.40} & \cellcolor{gray!15}\textbf{60.53} & \cellcolor{gray!15}\textbf{62.69} & \cellcolor{gray!15}\textbf{49.30} & \cellcolor{gray!15}\textbf{71.58} & \cellcolor{gray!15}\textbf{55.99} & \cellcolor{gray!15}\textbf{58.96} & \cellcolor{gray!15}\textbf{49.72} & \cellcolor{gray!15}\textbf{64.69} & \cellcolor{gray!15}\textbf{45.32} & \cellcolor{gray!15}\textbf{53.24} & \cellcolor{gray!15}36.48 & \cellcolor{gray!15}\textbf{62.78} & \cellcolor{gray!15}\textbf{46.50} & \cellcolor{gray!15}\textbf{48.59} \\ 
    \hline
    \rowcolor{TSoursblue}
    {} & \TSnoft & \cellcolor{gray!15}40.93 & \cellcolor{gray!15}9.54 & \cellcolor{gray!15}17.35 & \cellcolor{gray!15}22.60 & \cellcolor{gray!15}38.39 & \cellcolor{gray!15}9.56 & \cellcolor{gray!15}20.03 & \cellcolor{gray!15}22.66 & \cellcolor{gray!15}37.70 & \cellcolor{gray!15}8.99 & \cellcolor{gray!15}15.82 & \cellcolor{gray!15}20.84 & \cellcolor{gray!15}\textbf{41.96} & \cellcolor{gray!15}2.47 & \cellcolor{gray!15}23.06 & \cellcolor{gray!15}22.50 \\ 
    \rowcolor{TSoursblue}
    \multirow{-2}{*}{Qwen3-VL-8B} & \TSyesft & \cellcolor{gray!15}\textbf{54.63} & \cellcolor{gray!15}\textbf{76.21} & \cellcolor{gray!15}\textbf{61.37} & \cellcolor{gray!15}\textbf{64.07} & \cellcolor{gray!15}\textbf{50.24} & \cellcolor{gray!15}\textbf{72.02} & \cellcolor{gray!15}\textbf{55.83} & \cellcolor{gray!15}\textbf{59.37} & \cellcolor{gray!15}\textbf{50.07} & \cellcolor{gray!15}\textbf{64.12} & \cellcolor{gray!15}\textbf{44.77} & \cellcolor{gray!15}\textbf{52.99} & \cellcolor{gray!15}35.28 & \cellcolor{gray!15}\textbf{63.06} & \cellcolor{gray!15}\textbf{48.55} & \cellcolor{gray!15}\textbf{48.96} \\ 
    \hline
    \rowcolor{TSoursblue}
    {} & \TSnoft & \cellcolor{gray!15}19.20 & \cellcolor{gray!15}47.85 & \cellcolor{gray!15}35.71 & \cellcolor{gray!15}34.25 & \cellcolor{gray!15}20.86 & \cellcolor{gray!15}49.10 & \cellcolor{gray!15}39.90 & \cellcolor{gray!15}36.62 & \cellcolor{gray!15}21.76 & \cellcolor{gray!15}47.81 & \cellcolor{gray!15}40.81 & \cellcolor{gray!15}36.79 & \cellcolor{gray!15}14.84 & \cellcolor{gray!15}36.68 & \cellcolor{gray!15}44.22 & \cellcolor{gray!15}31.91 \\ 
    \rowcolor{TSoursblue}
    \multirow{-2}{*}{Qwen3.5-4B} & \TSyesft & \cellcolor{gray!15}\textbf{52.87} & \cellcolor{gray!15}\textbf{75.15} & \cellcolor{gray!15}\textbf{60.79} & \cellcolor{gray!15}\textbf{62.94} & \cellcolor{gray!15}\textbf{49.35} & \cellcolor{gray!15}\textbf{72.27} & \cellcolor{gray!15}\textbf{56.27} & \cellcolor{gray!15}\textbf{59.30} & \cellcolor{gray!15}\textbf{50.43} & \cellcolor{gray!15}\textbf{64.75} & \cellcolor{gray!15}\textbf{45.15} & \cellcolor{gray!15}\textbf{53.44} & \cellcolor{gray!15}\textbf{35.76} & \cellcolor{gray!15}\textbf{60.06} & \cellcolor{gray!15}\textbf{45.61} & \cellcolor{gray!15}\textbf{47.14} \\ 
    \hline
    \rowcolor{TSoursblue}
    {} & \TSnoft & \cellcolor{gray!15}40.10 & \cellcolor{gray!15}11.65 & \cellcolor{gray!15}17.42 & \cellcolor{gray!15}23.06 & \cellcolor{gray!15}38.38 & \cellcolor{gray!15}26.90 & \cellcolor{gray!15}21.54 & \cellcolor{gray!15}28.94 & \cellcolor{gray!15}40.71 & \cellcolor{gray!15}26.86 & \cellcolor{gray!15}18.92 & \cellcolor{gray!15}28.83 & \cellcolor{gray!15}\textbf{37.10} & \cellcolor{gray!15}25.85 & \cellcolor{gray!15}21.87 & \cellcolor{gray!15}28.27 \\ 
    \rowcolor{TSoursblue}
    \multirow{-2}{*}{Qwen3.5-9B} & \TSyesft & \cellcolor{gray!15}\textbf{50.50} & \cellcolor{gray!15}\textbf{72.79} & \cellcolor{gray!15}\textbf{60.08} & \cellcolor{gray!15}\textbf{61.13} & \cellcolor{gray!15}\textbf{48.24} & \cellcolor{gray!15}\textbf{71.29} & \cellcolor{gray!15}\textbf{56.90} & \cellcolor{gray!15}\textbf{58.81} & \cellcolor{gray!15}\textbf{50.56} & \cellcolor{gray!15}\textbf{66.17} & \cellcolor{gray!15}\textbf{47.58} & \cellcolor{gray!15}\textbf{54.77} & \cellcolor{gray!15}36.67 & \cellcolor{gray!15}\textbf{62.17} & \cellcolor{gray!15}\textbf{46.73} & \cellcolor{gray!15}\textbf{48.52} \\ 
    \noalign{\hrule height 1pt}
  \end{tabular}
  \end{adjustbox}
  \vspace{-0.1cm}
\caption{Proactive guidance results on ProactiveCoachBench and the EgoProactive validation set.}
  \vspace{-0.5cm}
  \label{tab:main_results}
  \label{tab:main_multigranular}
  \label{tab:main_multigranular_action}
  \label{tab:main_multigranular_step}
  \label{tab:existing_step}
\end{table*}


\subsection{Main Results}
\label{sec:main-results}

As shown in Table~\ref{tab:main_results}, our method consistently achieves higher overall performance than the baselines across all guidance levels on ProactiveCoachBench.
A higher sF1 indicates that the generated guidance is semantically well aligned with and sufficiently covers the ground-truth guidance.
A higher gF1 indicates that the model more accurately determines when to provide guidance and when to remain silent.
A higher PQS indicates better overall performance in both response--silence decisions and guidance quality.
Our method shows strong performance across all metrics, with consistent improvements across different backbones.

Our method also shows strong performance across all metrics on the EgoProactive validation set.
Some base models achieve relatively high sF1 but very low gF1, indicating that they tend to respond even when no guidance is needed.
Conversely, some models show relatively low sF1 but high PQS, indicating that they tend to remain silent even when a response is required.
Our method consistently achieves higher Avg. scores than the base models across various backbones.

\subsection{Ablation Studies and Analysis}
\label{sec:ablation-analysis}

\paragraph{Hierarchical vs. Step-Only Supervision.}
Table~\ref{tab:multilevel-supervision} compares step-only supervision with hierarchical supervision that jointly learns phase-, step-, and action-level guidance across different backbones.
Hierarchical supervision consistently improves sF1, gF1, PQS, and Avg. across all backbones.
These results show that learning hierarchically structured guidance not only enables guidance at multiple levels but also improves step-level proactive assistance.

\paragraph{Effect of Phase- and Action-Level Supervision.}
Table~\ref{tab:training-levels-ablation} compares step-only, phase--step, step--action, and phase--step--action supervision to analyze the effect of additional guidance levels.
Adding either phase- or action-level supervision improves step-level performance over step-only supervision, and using all three levels achieves the highest performance.
These results suggest that action-level supervision helps the model understand what actions make up each step and how they are progressing, providing useful evidence for determining whether the step is complete.
Phase-level supervision provides complementary context about the purpose and expected outcome of the current step and how it contributes to the overall procedure.
These results support the benefit of learning hierarchically structured guidance for step-level proactive assistance.

\begin{table}[t]
  \renewcommand{\cellcolor}[1]{}
  \centering
  \newsavebox{\PCtablethree}
  \newsavebox{\PCtablefour}
  \newsavebox{\PCtablefive}
  \newsavebox{\PCtablesix}
  \begin{lrbox}{\PCtablethree}
    \begin{minipage}[t]{0.49\linewidth}
      \vspace{0pt}  \centering
  \fontsize{8}{10.5}\selectfont
  \setlength{\tabcolsep}{3pt}
  \renewcommand{\arraystretch}{1.15}
  \begin{adjustbox}{max width=\linewidth}
  \begin{tabular}{@{}l@{\hspace{3pt}}l|ccc|c@{}} 
    \noalign{\hrule height 1pt}
    \multicolumn{2}{l|}{\multirow{2}{*}{Method}} & \multicolumn{4}{c}{ProactiveCoachBench Step} \\ 
    \cline{3-6} 
    & & sF1 & gF1 & PQS & Avg. \\ 
    \noalign{\hrule height 0.6pt}
    \multicolumn{6}{@{}l}{\textit{Qwen3-VL-4B}} \\ 
    {} & Step-only & \cellcolor{gray!15}42.85 & \cellcolor{gray!15}63.34 & \cellcolor{gray!15}41.78 & \cellcolor{gray!15}49.32 \\ 
    {} & Multi-level (Ours) & \cellcolor{gray!15}\textbf{49.30} & \cellcolor{gray!15}\textbf{71.58} & \cellcolor{gray!15}\textbf{55.99} & \cellcolor{gray!15}\textbf{58.96} \\ 
    \hline
    \multicolumn{6}{@{}l}{\textit{Qwen3-VL-8B}} \\ 
    {} & Step-only & \cellcolor{gray!15}45.82 & \cellcolor{gray!15}66.12 & \cellcolor{gray!15}47.27 & \cellcolor{gray!15}53.07 \\ 
    {} & Multi-level (Ours) & \cellcolor{gray!15}\textbf{50.24} & \cellcolor{gray!15}\textbf{72.02} & \cellcolor{gray!15}\textbf{55.83} & \cellcolor{gray!15}\textbf{59.37} \\ 
    \hline
    \multicolumn{6}{@{}l}{\textit{Qwen3.5-4B}} \\ 
    {} & Step-only & \cellcolor{gray!15}41.01 & \cellcolor{gray!15}64.53 & \cellcolor{gray!15}44.62 & \cellcolor{gray!15}50.06 \\ 
    {} & Multi-level (Ours) & \cellcolor{gray!15}\textbf{49.35} & \cellcolor{gray!15}\textbf{72.27} & \cellcolor{gray!15}\textbf{56.27} & \cellcolor{gray!15}\textbf{59.30} \\ 
    \hline
    \multicolumn{6}{@{}l}{\textit{Qwen3.5-9B}} \\ 
    {} & Step-only & \cellcolor{gray!15}42.56 & \cellcolor{gray!15}65.21 & \cellcolor{gray!15}45.77 & \cellcolor{gray!15}51.18 \\ 
    {} & Multi-level (Ours) & \cellcolor{gray!15}\textbf{48.24} & \cellcolor{gray!15}\textbf{71.29} & \cellcolor{gray!15}\textbf{56.90} & \cellcolor{gray!15}\textbf{58.81} \\ 
    \noalign{\hrule height 1pt}
  \end{tabular}
  \end{adjustbox}
  \vspace{-0.1cm}
  \caption{Step-only vs. hierarchical supervision.}
  \vspace{-0.1cm}
  \label{tab:multilevel-supervision}

    \end{minipage}
  \end{lrbox}
  \begin{lrbox}{\PCtablefour}
    \begin{minipage}[t]{0.49\linewidth}
      \vspace{0pt}  \centering
  \fontsize{8}{10.5}\selectfont
  \setlength{\tabcolsep}{3pt}
  \renewcommand{\arraystretch}{1.15}
  \begin{adjustbox}{max width=\linewidth}
  \begin{tabular}{@{}l@{\hspace{3pt}}l|ccc|c@{}} 
    \noalign{\hrule height 1pt}
    \multicolumn{2}{l|}{\multirow{2}{*}{Method}} & \multicolumn{4}{c}{ProactiveCoachBench Step} \\ 
    \cline{3-6} 
    & & sF1 & gF1 & PQS & Avg. \\ 
    \noalign{\hrule height 0.6pt}
    {} & Step-only & \cellcolor{gray!15}41.01 & \cellcolor{gray!15}64.53 & \cellcolor{gray!15}44.62 & \cellcolor{gray!15}50.06 \\ 
    {} & Phase-Step & \cellcolor{gray!15}48.21 & \cellcolor{gray!15}66.22 & \cellcolor{gray!15}46.37 & \cellcolor{gray!15}53.60 \\ 
    {} & Step-Action & \cellcolor{gray!15}46.16 & \cellcolor{gray!15}70.81 & \cellcolor{gray!15}55.02 & \cellcolor{gray!15}57.33 \\ 
    {} & Phase-Step-Action & \cellcolor{gray!15}\textbf{49.35} & \cellcolor{gray!15}\textbf{72.27} & \cellcolor{gray!15}\textbf{56.27} & \cellcolor{gray!15}\textbf{59.30} \\ 
    \noalign{\hrule height 1pt}
  \end{tabular}
  \end{adjustbox}
  \vspace{-0.1cm}
  \caption{Ablation on supervision levels.}
  \label{tab:training-levels-ablation}

    \end{minipage}
  \end{lrbox}
  \begin{lrbox}{\PCtablefive}
    \begin{minipage}[t]{\linewidth}
      \vspace{0pt}
\begingroup
  \centering
  \fontsize{8}{10.5}\selectfont
  \setlength{\tabcolsep}{3pt}
  \renewcommand{\arraystretch}{1.15}
  \begin{adjustbox}{max width=\linewidth}
  \begin{tabular}{@{}c|cccc|cccc|cccc@{}}
    \noalign{\hrule height 1pt}
    \multirow{2}{*}{FT} & \multicolumn{4}{c|}{Phase} & \multicolumn{4}{c|}{Step} & \multicolumn{4}{c}{Action} \\
    \cline{2-5}\cline{6-9}\cline{10-13}
    & sF1 & gF1 & PQS & Avg. & sF1 & gF1 & PQS & Avg. & sF1 & gF1 & PQS & Avg. \\
    \noalign{\hrule height 0.6pt}
    \TSnoft & 8.53 & 32.52 & 46.92 & 29.32 & 7.40 & 29.23 & 44.33 & 26.99 & 38.44 & 11.10 & 15.67 & 21.74 \\ 
    \TSyesft & \textbf{52.87} & \textbf{75.15} & \textbf{60.79} & \textbf{62.94} & \textbf{49.35} & \textbf{72.27} & \textbf{56.27} & \textbf{59.30} & \textbf{50.43} & \textbf{64.75} & \textbf{45.15} & \textbf{53.44} \\ 
    \noalign{\hrule height 1pt}
  \end{tabular}
  \end{adjustbox}
  \vspace{-0.1cm}
  \caption{Zero-shot multi-level guidance.}
  \vspace{-0.1cm}
  \label{tab:joint-hierarchical-guidance}
\endgroup

    \end{minipage}
  \end{lrbox}
  \begin{lrbox}{\PCtablesix}
    \begin{minipage}[t]{0.49\linewidth}
      \vspace{0pt}  \centering
  \fontsize{8}{10.5}\selectfont
  \setlength{\tabcolsep}{3pt}
  \renewcommand{\arraystretch}{1.15}
  \begin{adjustbox}{max width=\linewidth}
  \begin{tabular}{@{}ll|cccc@{}}
    \noalign{\hrule height 1pt}
    Transition & FT & sF1 & gF1 & PQS & Avg. \\
    \noalign{\hrule height 0.6pt}
    \multirow{2}{*}{Step$\rightarrow$Phase} & \TSnoft & \cellcolor{gray!15}24.18 & \cellcolor{gray!15}1.52 & \cellcolor{gray!15}2.69 & \cellcolor{gray!15}9.46 \\
     & \TSyesft & \cellcolor{gray!15}\textbf{67.66} & \cellcolor{gray!15}\textbf{93.78} & \cellcolor{gray!15}\textbf{95.17} & \cellcolor{gray!15}\textbf{85.54} \\
    \hline
    \multirow{2}{*}{Step$\rightarrow$Action} & \TSnoft & \cellcolor{gray!15}34.24 & \cellcolor{gray!15}8.43 & \cellcolor{gray!15}13.28 & \cellcolor{gray!15}18.65 \\
     & \TSyesft & \cellcolor{gray!15}\textbf{54.14} & \cellcolor{gray!15}\textbf{67.95} & \cellcolor{gray!15}\textbf{51.69} & \cellcolor{gray!15}\textbf{57.93} \\
    \hline
    \multirow{2}{*}{Phase$\rightarrow$Step} & \TSnoft & \cellcolor{gray!15}15.89 & \cellcolor{gray!15}5.50 & \cellcolor{gray!15}4.07 & \cellcolor{gray!15}8.49 \\
     & \TSyesft & \cellcolor{gray!15}\textbf{47.89} & \cellcolor{gray!15}\textbf{72.93} & \cellcolor{gray!15}\textbf{78.86} & \cellcolor{gray!15}\textbf{66.56} \\
    \hline
    \multirow{2}{*}{Action$\rightarrow$Step} & \TSnoft & \cellcolor{gray!15}38.06 & \cellcolor{gray!15}9.73 & \cellcolor{gray!15}8.32 & \cellcolor{gray!15}18.70 \\
     & \TSyesft & \cellcolor{gray!15}\textbf{59.98} & \cellcolor{gray!15}\textbf{82.68} & \cellcolor{gray!15}\textbf{78.45} & \cellcolor{gray!15}\textbf{73.70} \\
    \noalign{\hrule height 1pt}
  \end{tabular}
  \end{adjustbox}
  \vspace{-0.1cm}
  \caption{Guidance-level adaptation.}
  \label{tab:guidance-level-adaptation}

    \end{minipage}
  \end{lrbox}
  \usebox{\PCtablethree}\hfill
  \begin{minipage}[t]{0.49\linewidth}
    \vspace{0pt}\usebox{\PCtablefour}
    \par\vspace{6pt}
    \usebox{\PCtablesix}
  \end{minipage}
  \par\vspace{8pt}
  \usebox{\PCtablefive}
\end{table}

\begin{wrapfigure}[23]{r}{0.46\textwidth}
  \vspace{-0.3cm}
  \centering
  \includegraphics[pagebox=cropbox,width=0.96\linewidth]{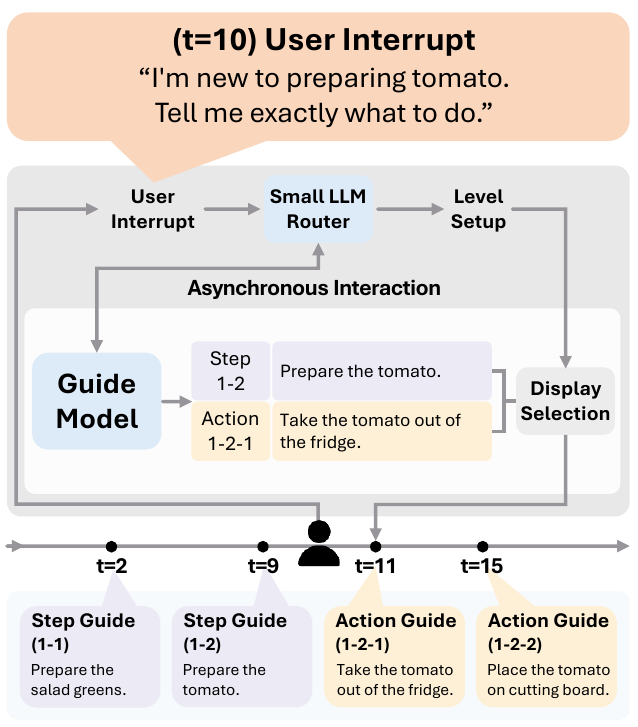}
  \caption{Guidance-level adaptation with a lightweight router. ProactiveCoach generates multi-level guidance, while the router selects the level requested by the user.}
  \label{fig:guidance-level-adaptation}
\end{wrapfigure}

\paragraph{Zero-Shot Multi-Level Guidance.}
We examine whether a base model can perform multi-level proactive guidance through prompting alone, without fine-tuning on hierarchically structured guidance.
Table~\ref{tab:joint-hierarchical-guidance} compares zero-shot Qwen3.5-4B with our fine-tuned model under the same multi-level prompt and three-level GT history.
Fine-tuning substantially improves performance across all three guidance levels.
These results suggest that prompting alone is insufficient for reliable multi-level proactive guidance, while training on hierarchically structured guidance helps the model learn the relationships across procedural levels and provide appropriate guidance at each level.

\paragraph{Guidance-Level Adaptation.}
\label{sec:async-level-adaptation}
A practical proactive procedural assistant should adapt its guidance level when the user's request changes during an ongoing task.
As illustrated in Figure~\ref{fig:guidance-level-adaptation}, our system separates guidance generation from guidance-level selection.
Our trained model generates guidance at the three levels, while a lightweight LLM router selects which level to present based on the user's request.
Because the two modules operate asynchronously and independently, the displayed guidance level can change without modifying or fine-tuning the guide model.

We evaluate this on four level transitions: Step$\rightarrow$Phase, Step$\rightarrow$Action, Phase$\rightarrow$Step, and Action$\rightarrow$Step.
For the baseline, the level-change request is added to the dialogue history, and the model is prompted to adjust its guidance level accordingly.
In our system, the router handles the level change while the guide model continues to generate multi-level guidance.
As shown in Table~\ref{tab:guidance-level-adaptation}, our system improves overall performance by 57.1\%p on average across the four transitions compared with the in-context adaptation baseline.
For the baseline, the previous dialogue history contains guidance consistently provided at the original level, making it difficult to adapt to a new level through in-context prompting alone.
In contrast, separating guidance generation from level selection allows our system to change the displayed guidance level without being constrained by the previous guidance history.
These results show that our design can reliably adapt guidance granularity to the user's request without additional fine-tuning of the guide model.

\section{Conclusion}
In this work, we explored hierarchical procedural understanding for proactive AI assistance.
Existing datasets either remain limited to detection-based proactive understanding or provide procedural guidance at a fixed granularity, limiting supervision for tracking task progress and adapting guidance granularity.
We introduced ProactiveCoach-Instruct for training and ProactiveCoachBench for evaluation, providing hierarchically structured guidance at the phase, step, and action levels.
We further proposed ProactiveCoach, a hierarchical guidance learning method that jointly learns multi-level proactive guidance in a single vision-language model.
Our method consistently improves proactive assistance across different backbones, and our adaptive guidance system supports guidance-level changes without additional fine-tuning of the guide model.
We believe that our proposed datasets and method provide a useful foundation for future research on proactive AI assistance.

\bibliography{iclr2027_conference}

@article{pwr,
  title={Plan, Watch, Recover: A Benchmark and Architectures for Proactive Procedural Assistance},
  author={Kundu, Kaustav and Shrivastava, Ritvik and Arap, Maxim and Wang, Nanshu and Zhu, Xianhui and Fettes, Quintin and Tiwari, Gautam and Suresh, Parth and Moutakanni, Th{\'e}o and Munoz, Alejandro Castillejo and others},
  journal={arXiv preprint arXiv:2606.04970},
  year={2026}
}

@inproceedings{guideme,
  title={GuideMe: Multi-domain Task Guidance and Intervention in Streaming Video},
  author={Liu, Fang and Chen, Jinpeng and Xu, Ke and Liu, Yuhao and Guan, Huankang and Lu, Xudong and Bo, Yang and Hancke, Gerhard and Liu, Rui and Lau, Rynson WH},
  booktitle={European Conference on Computer Vision},
  pages={57--74},
  year={2026},
  organization={Springer}
}

@inproceedings{proassist,
  title={Proactive Assistant Dialogue Generation from Streaming Egocentric Videos},
  author={Zhang, Yichi and Dong, Xin Luna and Lin, Zhaojiang and Madotto, Andrea and Kumar, Anuj and Damavandi, Babak and Chai, Joyce and Moon, Seungwhan},
  booktitle={Proceedings of the 2025 Conference on Empirical Methods in Natural Language Processing},
  pages={12044--12068},
  year={2025}
}

@article{streampro,
  title={StreamPro: From Reactive Perception to Proactive Decision-Making in Streaming Video},
  author={Li, Ao and Xiao, Zihan and Yue, Zihao and Xu, Boshen and Yao, Linli and Li, Jiaze and Fu, Pei and Ju, Jianzhong and Luan, Jian and Jin, Qin},
  journal={arXiv preprint arXiv:2605.16381},
  year={2026}
}

@article{videochat3,
  title={VideoChat3: Fully Open Video MLLM for Efficient and Generalist Video Understanding},
  author={Li, Xinhao and Zhu, Yuhan and Zeng, Xiangyu and Dong, Yuhao and Wu, Haoning and Zhang, Zhiqiu and Yang, Yuandong and Ma, Changlian and Zhang, Qingyu and Shi, Yansong and others},
  journal={arXiv preprint arXiv:2607.14935},
  year={2026}
}

@inproceedings{mmduet2,
  title={Mmduet2: Enhancing proactive interaction of video mllms with multi-turn reinforcement learning},
  author={Wang, Yueqian and Liu, Songxiang and Wang, Disong and Xu, Nuo and Wan, Guanglu and Zhang, Huishuai and Zhao, Dongyan},
  booktitle={International Conference on Learning Representations},
  volume={2026},
  pages={23257--23270},
  year={2026}
}

@inproceedings{streamo,
  title={Streaming video instruction tuning},
  author={Xia, Jiaer and Chen, Peixian and Zhang, Mengdan and Sun, Xing and Zhou, Kaiyang},
  booktitle={Proceedings of the IEEE/CVF Conference on Computer Vision and Pattern Recognition},
  pages={31219--31229},
  year={2026}
}

@article{em-grade,
  title={Em-garde: A propose-match framework for proactive streaming video understanding},
  author={Zheng, Yikai and Ding, Xin and Yang, Yifan and Jiang, Shiqi and Wu, Hao and Zhang, Qianxi and Wang, Weijun and Cao, Ting and Liu, Yunxin},
  journal={arXiv preprint arXiv:2603.19054},
  year={2026}
}

@article{omnipro,
  title={OmniPro: A Comprehensive Benchmark for Omni-Proactive Streaming Video Understanding},
  author={Zhao, Ruixiang and Yang, Jie and Xin, Zijie and Wang, Tianyi and Rao, Fengyun and LYU, Jing and Li, Xirong},
  journal={arXiv preprint arXiv:2605.18577},
  year={2026}
}

@inproceedings{streamready,
  title={Streamready: Learning what to answer and when in long streaming videos},
  author={Azad, Shehreen and Vineet, Vibhav and Rawat, Yogesh S},
  booktitle={Proceedings of the IEEE/CVF Conference on Computer Vision and Pattern Recognition},
  pages={40494--40504},
  year={2026}
}

@inproceedings{querystream,
  title={Querystream: Advancing streaming video understanding with query-aware pruning and proactive response},
  author={Zhang, Kairui and Yang, Zhenyu and Wang, Bing and Qian, Shengsheng and Xu, Changsheng},
  booktitle={The Fourteenth International Conference on Learning Representations},
  year={2026}
}

@article{proact-vl,
  title={Proact-VL: A Proactive VideoLLM for Real-Time AI Companions},
  author={Yan, Weicai and Dai, Yuhong and Ran, Qi and Li, Haodong and Lin, Wang and Jin, Tao and Xie, Xing and Liao, Hao and Lian, Jianxun},
  journal={arXiv preprint arXiv:2603.03447},
  year={2026}
}

@inproceedings{response-g1,
  title={Response-G1: Explicit Scene Graph Modeling for Proactive Streaming Video Understanding},
  author={Ma, Ke and Tang, Jiaqi and Guo, Bin and Han, Xueting and Xu, Ruonan and He, Qingfeng and Wang, Ziheng and Wang, Xu and Chen, Qifeng and Yu, Zhiwen and others},
  booktitle={Proceedings of the 64th Annual Meeting of the Association for Computational Linguistics (Volume 1: Long Papers)},
  pages={44139--44153},
  year={2026}
}

@article{eyewo,
  title={Eyes Wide Open: Ego Proactive Video-LLM for Streaming Video},
  author={Yu, Xueyang and Shi, Cheng and Wang, Yang and Yang, Sibei},
  journal={Advances in Neural Information Processing Systems},
  volume={38},
  pages={13420--13463},
  year={2026}
}

@article{proactivevideoqa,
  title={Proactivevideoqa: A comprehensive benchmark evaluating proactive interactions in video large language models},
  author={Wang, Yueqian and Meng, Xiaojun and Wang, Yifan and Zhang, Huishuai and Zhao, Dongyan},
  journal={arXiv preprint arXiv:2507.09313},
  year={2025}
}

@article{ipibench,
  title={IPIBench: Evaluating Interactive Proactive Intelligence of MLLMs under Continuous Streams},
  author={Li, Jinzhao and Chen, Yinuo and Song, Wenxuan and Lei, Yijia and Zhang, Yichi and Yan, Honglei and Pan, Panwang and Liu, Miao},
  journal={arXiv preprint arXiv:2605.27074},
  year={2026}
}

@inproceedings{streamingbench,
  title={Streamingbench: Assessing the gap for mllms to achieve streaming video understanding},
  author={Lin, Junming and Fang, Zheng and Chen, Chi and Cheng, Haoxuan and Wan, Zihao and Luo, Fuwen and Wang, Ziyue and Li, Peng and Liu, Yang and Sun, Maosong},
  booktitle={ICASSP 2026-2026 IEEE International Conference on Acoustics, Speech and Signal Processing (ICASSP)},
  pages={12147--12151},
  year={2026},
  organization={IEEE}
}

@misc{ovobench,
      title={OVO-Bench: How Far is Your Video-LLMs from Real-World Online Video Understanding?}, 
      author={Yifei Li and Junbo Niu and Ziyang Miao and Chunjiang Ge and Yuanhang Zhou and Qihao He and Xiaoyi Dong and Haodong Duan and Shuangrui Ding and Rui Qian and Pan Zhang and Yuhang Zang and Yuhang Cao and Conghui He and Jiaqi Wang},
      year={2025},
      eprint={2501.05510},
      archivePrefix={arXiv},
      primaryClass={cs.CV},
      url={https://arxiv.org/abs/2501.05510}, 
}

@inproceedings{videorecap,
  title={Video recap: Recursive captioning of hour-long videos},
  author={Islam, Md Mohaiminul and Ho, Ngan and Yang, Xitong and Nagarajan, Tushar and Torresani, Lorenzo and Bertasius, Gedas},
  booktitle={2024 IEEE/CVF Conference on Computer Vision and Pattern Recognition (CVPR)},
  pages={18198--18208},
  year={2024},
  organization={IEEE}
}

@inproceedings{openhsvu,
  title={Open-ended hierarchical streaming video understanding with vision language models},
  author={Kang, Hyolim and Park, Yunsu and Yoo, Youngbeom and Choi, Yeeun and Kim, Seon Joo},
  booktitle={2025 IEEE/CVF International Conference on Computer Vision (ICCV)},
  pages={20715--20725},
  year={2025},
  organization={IEEE}
}

@inproceedings{hirest,
  title={Hierarchical video-moment retrieval and step-captioning},
  author={Zala, Abhay and Cho, Jaemin and Kottur, Satwik and Chen, Xilun and Oguz, Barlas and Mehdad, Yashar and Bansal, Mohit},
  booktitle={2023 IEEE/CVF Conference on Computer Vision and Pattern Recognition (CVPR)},
  pages={23056--23065},
  year={2023},
  organization={IEEE}
}

@article{guide,
  title={GUIDE: a guideline-guided dataset for instructional video comprehension},
  author={Liang, Jiafeng and Jiang, Shixin and Wang, Zekun and Pan, Haojie and Chen, Zerui and Chu, Zheng and Liu, Ming and Fu, Ruiji and Wang, Zhongyuan and Qin, Bing},
  journal={arXiv preprint arXiv:2406.18227},
  year={2024}
}

@article{action100m,
  title={Action100m: A large-scale video action dataset},
  author={Chen, Delong and Kasarla, Tejaswi and Bang, Yejin and Shukor, Mustafa and Chung, Willy and Yu, Jade and Bolourchi, Allen and Moutakanni, Theo and Fung, Pascale},
  journal={arXiv preprint arXiv:2601.10592},
  year={2026}
}

@article{goalstep,
  title={Ego4d goal-step: Toward hierarchical understanding of procedural activities},
  author={Song, Yale and Byrne, Eugene and Nagarajan, Tushar and Wang, Huiyu and Martin, Miguel and Torresani, Lorenzo},
  journal={Advances in neural information processing systems},
  volume={36},
  pages={38863--38886},
  year={2023}
}

@inproceedings{ego-exo4d,
  title={Ego-exo4d: Understanding skilled human activity from first-and third-person perspectives},
  author={Grauman, Kristen and Westbury, Andrew and Torresani, Lorenzo and Kitani, Kris and Malik, Jitendra and Afouras, Triantafyllos and Ashutosh, Kumar and Baiyya, Vijay and Bansal, Siddhant and Boote, Bikram and others},
  booktitle={Proceedings of the IEEE/CVF conference on computer vision and pattern recognition},
  pages={19383--19400},
  year={2024}
}

@inproceedings{epic-kitchens,
  title={Scaling Egocentric Vision: The EPIC-KITCHENS Dataset},
  author={Damen, Dima and Doughty, Hazel and Farinella, Giovanni Maria and Fidler, Sanja and Furnari, Antonino and Kazakos, Evangelos and Moltisanti, Davide and Munro, Jonathan and Perrett, Toby and Price, Will and others},
  booktitle={European conference on computer vision},
  pages={753--771},
  year={2018},
  organization={Springer}
}

@article{captaincook4d,
  title={Captaincook4d: A dataset for understanding errors in procedural activities},
  author={Peddi, Rohith and Arya, Shivvrat and Challa, Bharath and Pallapothula, Likhitha and Vyas, Akshay and Gouripeddi, Bhavya and Zhang, Qifan and Wang, Jikai and Komaragiri, Vasundhara and Ragan, Eric and others},
  journal={Advances in Neural Information Processing Systems},
  volume={37},
  pages={135626--135679},
  year={2024}
}

@inproceedings{egoexolearn,
  title={Egoexolearn: A dataset for bridging asynchronous ego-and exo-centric view of procedural activities in real world},
  author={Huang, Yifei and Chen, Guo and Xu, Jilan and Zhang, Mingfang and Yang, Lijin and Pei, Baoqi and Zhang, Hongjie and Dong, Lu and Wang, Yali and Wang, Limin and others},
  booktitle={2024 IEEE/CVF Conference on Computer Vision and Pattern Recognition (CVPR)},
  pages={22072--22086},
  year={2024},
  organization={IEEE}
}

@inproceedings{hd-epic,
  title={Hd-epic: A highly-detailed egocentric video dataset},
  author={Perrett, Toby and Darkhalil, Ahmad and Sinha, Saptarshi and Emara, Omar and Pollard, Sam and Parida, Kranti Kumar and Liu, Kaiting and Gatti, Prajwal and Bansal, Siddhant and Flanagan, Kevin and others},
  booktitle={2025 IEEE/CVF Conference on Computer Vision and Pattern Recognition (CVPR)},
  pages={23901--23913},
  year={2025},
  organization={IEEE}
}

@inproceedings{wtag,
  title={Can foundation models watch, talk and guide you step by step to make a cake?},
  author={Bao, Yuwei and Yu, Keunwoo and Zhang, Yichi and Storks, Shane and Bar-Yossef, Itamar and de la Iglesia, Alex and Su, Megan and Zheng, Xiao and Chai, Joyce},
  booktitle={Findings of the Association for Computational Linguistics: EMNLP 2023},
  pages={12325--12341},
  year={2023}
}

@inproceedings{holoassist,
  title={Holoassist: an egocentric human interaction dataset for interactive ai assistants in the real world},
  author={Wang, Xin and Kwon, Taein and Rad, Mahdi and Pan, Bowen and Chakraborty, Ishani and Andrist, Sean and Bohus, Dan and Feniello, Ashley and Tekin, Bugra and Frujeri, Felipe Vieira and others},
  booktitle={2023 IEEE/CVF International Conference on Computer Vision (ICCV)},
  pages={20213--20224},
  year={2023},
  organization={IEEE}
}

@article{gemini2.5,
  title={Gemini 2.5: Pushing the frontier with advanced reasoning, multimodality, long context, and next generation agentic capabilities},
  author={Comanici, Gheorghe and Bieber, Eric and Schaekermann, Mike and Pasupat, Ice and Sachdeva, Noveen and Dhillon, Inderjit and Blistein, Marcel and Ram, Ori and Zhang, Dan and Rosen, Evan and others},
  journal={arXiv preprint arXiv:2507.06261},
  year={2025}
}

@misc{claudehaiku,
  author       = {Anthropic},
  title        = {Claude Haiku 4.5},
  year         = {2025},
  url          = {https://www.anthropic.com/claude/haiku},
  note         = {Large language model}
}

@article{gpt5,
  title={Openai gpt-5 system card},
  author={Singh, Aaditya and Fry, Adam and Perelman, Adam and Tart, Adam and Ganesh, Adi and El-Kishky, Ahmed and McLaughlin, Aidan and Low, Aiden and Ostrow, AJ and Ananthram, Akhila and others},
  journal={arXiv preprint arXiv:2601.03267},
  year={2025}
}

@inproceedings{videollm-online,
  title={Videollm-online: Online video large language model for streaming video},
  author={Chen, Joya and Lv, Zhaoyang and Wu, Shiwei and Lin, Kevin Qinghong and Song, Chenan and Gao, Difei and Liu, Jia-Wei and Gao, Ziteng and Mao, Dongxing and Shou, Mike Zheng},
  booktitle={2024 IEEE/CVF Conference on Computer Vision and Pattern Recognition (CVPR)},
  pages={18407--18418},
  year={2024},
  organization={IEEE}
}

@article{videochat-r1.5,
  title={Videochat-r1. 5: Visual test-time scaling to reinforce multimodal reasoning by iterative perception},
  author={Yan, Ziang and He, Yinan and Li, Xinhao and Yue, Zhengrong and Zeng, Xiangyu and Wang, Yali and Qiao, Yu and Wang, Limin and Wang, Yi},
  journal={Advances in Neural Information Processing Systems},
  volume={38},
  pages={119152--119184},
  year={2026}
}

@article{internvl3,
  title={Internvl3: Exploring advanced training and test-time recipes for open-source multimodal models},
  author={Zhu, Jinguo and Wang, Weiyun and Chen, Zhe and Liu, Zhaoyang and Ye, Shenglong and Gu, Lixin and Tian, Hao and Duan, Yuchen and Su, Weijie and Shao, Jie and others},
  journal={arXiv preprint arXiv:2504.10479},
  year={2025}
}

@article{videollama3,
  title={Videollama 3: Frontier multimodal foundation models for image and video understanding},
  author={Zhang, Boqiang and Li, Kehan and Cheng, Zesen and Hu, Zhiqiang and Yuan, Yuqian and Chen, Guanzheng and Leng, Sicong and Jiang, Yuming and Zhang, Hang and Li, Xin and others},
  journal={arXiv preprint arXiv:2501.13106},
  year={2025}
}

@article{llava-video,
  title={Llava-video: Video instruction tuning with synthetic data},
  author={Zhang, Yuanhan and Wu, Jinming and Li, Wei and Li, Bo and Ma, Zejun and Liu, Ziwei and Li, Chunyuan},
  journal={arXiv preprint arXiv:2410.02713},
  year={2024}
}

@article{qwen3-vl,
  title={Qwen3-vl technical report},
  author={Bai, Shuai and Cai, Yuxuan and Chen, Ruizhe and Chen, Keqin and Chen, Xionghui and Cheng, Zesen and Deng, Lianghao and Ding, Wei and Gao, Chang and Ge, Chunjiang and others},
  journal={arXiv preprint arXiv:2511.21631},
  year={2025}
}

@article{qwen3.5,
  title={Qwen3. 5-omni technical report},
  author={Team, Qwen},
  journal={arXiv preprint arXiv:2604.15804},
  year={2026}
}
\bibliographystyle{iclr2027_conference}


\appendix
\input{sec/7_supp}

\end{document}